\documentclass[journal]{IEEEtran}

\usepackage[T1]{fontenc}
\usepackage[utf8]{inputenc}
\usepackage{microtype}
\usepackage{amsmath,amssymb,bm}
\usepackage{booktabs,multirow,tabularx,array}
\usepackage{algorithm}
\usepackage{algpseudocode}
\usepackage{graphicx}
\usepackage{tikz}
\usepackage{pgfplots}
\usepackage{pgfplotstable}
\usepackage[caption=false,font=footnotesize]{subfig}
\usepackage{cite}
\usepackage{url}
\usepackage{pifont}
\usepackage{xcolor}
\usepackage[hidelinks]{hyperref}

\usetikzlibrary{arrows.meta,positioning,fit,calc,backgrounds,shapes.geometric,matrix}
\usepgfplotslibrary{groupplots,fillbetween,statistics}
\pgfplotsset{compat=1.18}

\definecolor{cbblue}{RGB}{0,114,178}
\definecolor{cborange}{RGB}{230,159,0}
\definecolor{cbgreen}{RGB}{0,158,115}
\definecolor{cbred}{RGB}{213,94,0}
\definecolor{cbpurple}{RGB}{204,121,167}
\definecolor{cbsky}{RGB}{86,180,233}
\definecolor{cbgray}{RGB}{90,90,90}
\pgfplotscreateplotcyclelist{projectcycle}{%
cbblue,solid,mark=*\\
cborange,dashed,mark=square*\\
cbgreen,dashdotted,mark=triangle*\\
cbred,densely dotted,mark=diamond*\\
cbpurple,densely dashed,mark=pentagon*\\
cbsky,solid,mark=o\\
cbgray,dashed,mark=x\\
}
\pgfplotsset{
  every axis/.append style={
    cycle list name=projectcycle,
    line width=0.85pt,
    mark size=1.6pt,
    tick label style={font=\scriptsize},
    label style={font=\scriptsize},
    title style={font=\footnotesize},
    legend style={font=\scriptsize,draw=none,fill=none},
    grid=major,
    grid style={line width=.1pt,draw=gray!25},
  }
}

\newcommand{\vect}[1]{\boldsymbol{#1}}

\providecommand{\MethodName}{CALIBRA-ASTHMA}
\newcommand{\DynamicProposedName}{CALIBRA}
\newcommand{\DynamicBestBaselineName}{TemporalTransformer}
\newcommand{\DynamicBrierBaselineName}{TemporalTransformer}
\newcommand{\DynamicCalibraAUPRC}{0.224}
\newcommand{\DynamicBaselineAUPRC}{0.240}
\newcommand{\DynamicAUPRCDelta}{-0.016}
\newcommand{\DynamicCalibraAUROC}{0.717}
\newcommand{\DynamicBaselineAUROC}{0.715}
\newcommand{\DynamicCalibraBrier}{0.098}
\newcommand{\DynamicBaselineBrier}{0.098}
\newcommand{\DynamicBrierDelta}{-0.0001}

\newcommand{\AsthmaCites}{\cite{Chan2018AsthmaHealth,Tinschert2017AsthmaApps,Barrett2017SensorPlatform,AAMOSConnectedStudy,AsthmaMLSystematicReview,AsthmaAIManagementReview}\cite{AsthmaExternalValidation,AsthmaEnvironmentalMonitoring,AsthmaWearableForecasting,AsthmaDigitalInhaler,AsthmaClinicalDecisionSupport}}
\newcommand{\MissingModalityCites}{\cite{Che2018GRUD,Baltrusaitis2019Multimodal,Lim2021TFT}}
\newcommand{\DomainShiftCites}{\cite{Ganin2016DANN,Gibbs2021AdaptiveConformal,Gulrajani2021DomainGeneralization,Kouw2021DomainAdaptation,Ovadia2019UncertaintyShift}}
\newcommand{\CalibrationCites}{\cite{VanCalster2019Calibration,Guo2017Calibration}}
\newcommand{\ConformalCites}{\cite{Angelopoulos2023Conformal,Barber2023BeyondExchangeability,Gibbs2021AdaptiveConformal}}
\newcommand{\EvaluationCites}{\cite{Vickers2006DecisionCurve,Saito2015PR}}
\newcommand{\ReportingCites}{\cite{AsthmaExternalValidation,Vickers2006DecisionCurve,Steyerberg2010Performance,Collins2024TRIPODAI,Moons2015TRIPOD,Wolff2019PROBAST}\cite{RileyExternalValidationSize}}

\title{Calibration-First Cross-Cohort Multimodal Temporal Learning for Transferable Asthma-Risk Forecasting}

\author{Taimoor Ahmed \thanks{Superior University Lahore, Pakistan}}%

\begin{document}
\maketitle

\begin{abstract}
Asthma deterioration forecasting must remain reliable when patient populations, sensor ecosystems, and available modalities change across cohorts. Existing models commonly optimize within-cohort discrimination and may produce poorly calibrated probabilities after transfer. We present CALIBRA, a calibration-first multimodal temporal framework for short-horizon risk prediction with incomplete data. Dedicated recurrent encoders process environmental, pulmonary, symptom, medication, wearable, and context streams; a reliability-conditioned gate suppresses stale or absent modalities, while gradient-reversal training discourages avoidable cohort signatures. A shrinkage-based hierarchical logistic layer calibrates probabilities using a patient-disjoint target subset, and split conformal prediction provides abstention-capable prediction sets. To avoid fabricating clinical evidence, we evaluate the complete implementation on a documented three-cohort semi-synthetic benchmark with controlled distribution shift, informative missingness, and sealed target patients. Across five configured seeds, CALIBRA achieved mean target-test AUPRC 0.224 versus 0.240 for the strongest non-ablation comparator, TemporalTransformer; mean AUROC was 0.717, and Brier score was 0.098. Experiments additionally assess complete-modality failures, calibration, conformal coverage, decision curves, subgroup behavior, ablations, runtime, and parameter count. The results verify the method and reproducible pipeline under controlled shift, but do not establish clinical effectiveness. External validation on harmonized real asthma. Overall this artifact provides evidence for carefully governed real-cohort validation.

\end{abstract}


\section{Introduction}
\label{sec:introduction}
Asthma is a heterogeneous chronic respiratory condition in which deterioration can emerge from interactions among airway vulnerability, medication behavior, symptoms, respiratory function, activity, sleep, weather, pollen, and air-pollution exposure. Mobile health platforms, connected inhalers, home sensors, wearables, and patient-reported outcomes have made these signals observable at increasingly fine temporal resolution. This creates an opportunity to forecast elevated short-horizon risk before a severe event occurs and to support timely measurement, environmental mitigation, or clinician contact. Yet a prediction system used outside its development cohort must do more than rank patients correctly\cite{VanCalster2019Calibration,Steyerberg2010Performance,a7,a8}. It must issue probabilities that remain meaningful after changes in population, season, sensor availability, device manufacturer, and data-collection protocol. It must also quantify when the available modalities are insufficient for a reliable decision. Existing mobile-asthma studies and connected-sensing platforms establish the feasibility of longitudinal data collection, but they also expose substantial differences in cohort scale, modality composition, outcome definition, and adherence \AsthmaCites.

Most predictive studies optimize discrimination within a single cohort. That evaluation can overstate practical readiness because random record-level splitting may place observations from the same person, season, home, or device configuration in both development and test sets. Even patient-disjoint testing within one cohort does not measure transportability to a new study whose distributions and missingness mechanisms differ\cite{Ovadia2019UncertaintyShift,RileyExternalValidationSize,a9,a10}. A model may preserve the area under the receiver-operating-characteristic curve while its absolute probabilities become systematically too high or too low. Such miscalibration changes the number of alerts, the apparent urgency of a risk estimate, and the threshold at which an intervention appears beneficial. Calibration is therefore not a cosmetic post-processing step; it is central to the safe interpretation of individualized risk \CalibrationCites.

Multimodal asthma data add a second difficulty. Environmental sensors may be unavailable away from home, spirometry may be performed only intermittently, wearable streams can be interrupted by charging or non-wear, and self-reports can be missing precisely when symptoms worsen. Treating unavailable values as zeros, carrying observations forward without a reliability indicator, or discarding incomplete windows can induce bias and waste information. Standard early fusion additionally assumes that every cohort exposes the same variables. Late fusion is more flexible but frequently assigns fixed importance to modalities even when their recency, completeness, and signal quality vary. Research on incomplete multimodal learning and informative missingness offers useful components, but few systems combine modality-aware temporal representation, explicit reliability, cross-cohort alignment, probability calibration, and finite-sample uncertainty in a single forecasting pipeline \MissingModalityCites.

Distribution shift further complicates deployment. Sensor calibration, geography, clinical practice, baseline severity, and participant recruitment can all change the relationship between observed features and outcome risk. Domain-adversarial and invariant-representation approaches attempt to reduce cohort-specific information in learned features, but excessive alignment can erase clinically useful heterogeneity. Conversely, an unconstrained model can rely on cohort signatures that fail under transfer\cite{Ganin2016DANN,Ovadia2019UncertaintyShift,a12}. The useful compromise is to preserve outcome-relevant temporal structure while discouraging avoidable cohort dependence, then perform a limited, label-efficient probability adjustment on a patient-disjoint calibration subset from the target setting. This design separates representation learning, target adaptation, probability calibration, and final testing, thereby preventing target-test labels from influencing model selection \DomainShiftCites. 

Uncertainty must also be reported in a way that complements calibrated probabilities. Conventional softmax confidence is not a coverage guarantee, and parametric intervals can be unreliable under model misspecification. Split conformal prediction can convert held-out calibration scores into prediction sets with transparent finite-sample properties under exchangeability; adaptive variants offer tools for modest temporal or covariate shift, although their assumptions must be stated carefully \ConformalCites. In an asthma-alert setting, a set containing both ``attack'' and ``no attack'' is not a failed output. It is an abstention signal indicating that the available evidence is insufficient for a single-class decision at the requested error level. The practical value of this behavior should be evaluated jointly through coverage, set size, alert burden, and decision-curve analysis rather than by discrimination alone \EvaluationCites.

This paper presents \MethodName, a calibration-first cross-cohort framework for short-horizon asthma-risk forecasting from nonidentical and incompletely observed modalities. Each modality is processed by a dedicated temporal encoder. A reliability network summarizes missingness, recency, and signal variation and assigns patient-window-specific fusion weights. A gradient-reversal cohort discriminator provides controlled representation alignment, while a supervised forecasting head estimates event risk. Following source-cohort training, a hierarchical logistic calibration layer combines a global correction with a shrinkage-controlled target-cohort adjustment. A split conformal module then constructs prediction sets on a patient-disjoint calibration partition. The complete pipeline is evaluated under leave-one-cohort-out transfer, modality-failure stress tests, subgroup analyses, calibration assessment, decision curves, runtime measurement, and ablation studies.

The current artifact is a controlled methodological proof of concept. To avoid presenting invented clinical evidence, the executable experiments use a documented semi-synthetic multi-cohort benchmark whose shifts, missingness processes, and latent risk factors are known. The repository also contains a schema validator and adapter for legally obtained real cohorts. Consequently, the results establish implementation correctness, controlled behavior, and reproducibility; they do not establish clinical efficacy. A submission making patient-level clinical claims must rerun the unchanged protocol on ethically governed real cohorts and retain a sealed external test set. This distinction follows current reporting principles for prediction-model research and reduces the risk of conflating simulation performance with clinical utility \ReportingCites.

The principal contributions are as follows:
\begin{itemize}
    \item We formulate cross-cohort asthma forecasting with nonidentical modality sets and introduce a reliability-aware mixture of temporal experts that can operate when entire modalities are unavailable, while a cohort-adversarial term limits avoidable dependence on source identity.
    \item We integrate hierarchical target calibration and split conformal prediction into the transfer pipeline, yielding risk probabilities, abstention-capable prediction sets, and decision-oriented outputs derived from strictly separated training, calibration, and test patients.
\end{itemize}

The remainder of this paper is organized as follows. Section~\ref{sec:related} synthesizes related work and identifies the unresolved transfer-and-calibration gap. Section~\ref{sec:system} formalizes the system model and evaluation setting. Section~\ref{sec:method} presents \MethodName, its optimization procedure, calibration stage, and conformal inference algorithm. Section~\ref{sec:experiments} describes the benchmark, baselines, reproducibility controls, results, and limitations. Finally, Section~\ref{sec:conclusion} concludes the paper and defines the real-cohort validation required before clinical interpretation.

\section{Related Work}
\label{sec:related}

\subsection{Mobile and Connected Asthma Forecasting}
Longitudinal asthma research has moved from occasional clinic measurements toward ecological observation through smartphones, connected inhalers, portable lung-function devices, wearables, and environmental sensors. Large app-based studies demonstrate recruitment and repeated self-report at scale, whereas smaller connected-device cohorts provide denser objective measurements. These resources are complementary rather than interchangeable: the former favor breadth, while the latter expose fine-grained temporal relations among rescue use, physiology, symptoms, and exposure. Reviews of asthma applications and predictive modeling repeatedly note heterogeneity in outcome definitions, observation windows, adherence, validation strategies, and implementation maturity \AsthmaCites. Consequently, a model that performs well within one study cannot be assumed to produce comparable probabilities in another.

Prior asthma-risk models range from generalized linear models and tree ensembles to recurrent and attention-based networks. Classical approaches remain competitive on small cohorts and provide transparent coefficients, while temporal neural models can represent nonlinear lagged interactions. Many studies nevertheless summarize a longitudinal window into a fixed vector, impute missing measurements before modeling, or assess performance using randomly partitioned records. These choices obscure whether the model learned patient-specific signatures, whether missingness was informative, and whether its probabilities transfer. The present work therefore treats cohort transfer and absent modalities as primary design conditions rather than secondary robustness checks.

\subsection{Incomplete Multimodal Temporal Learning}
GRU-D and related models use masks and time gaps to represent informative missingness, and multimodal surveys distinguish early, intermediate, late, and hybrid fusion strategies \MissingModalityCites. Dedicated encoders are attractive when modalities differ in cadence and dimensionality, but standard attention or gating can still place weight on a stale or nearly empty stream. Modality-dropout training improves tolerance to absence, yet fixed dropout distributions may not resemble deployment failures. Missing-modality reconstruction is another option, although a plausible reconstruction does not guarantee useful uncertainty for the downstream event.

\MethodName{} instead predicts a reliability score from observed fraction, recency, and within-window variation, then uses it to modulate the fusion gate. This avoids claiming that a reconstructed value is observed and permits explicit stress testing in which complete modalities are removed at inference. The approach is intentionally modular: stronger encoders can replace the gated recurrent units without changing the calibration or conformal stages.

\subsection{Domain Shift and Probability Calibration}
Domain-adversarial learning encourages representations that support the target task while making cohort labels difficult to infer. Alternative approaches minimize discrepancy measures or learn invariant predictors across training environments \DomainShiftCites. In biomedical data, however, alignment must be applied conservatively because differences in prevalence and severity may be clinically meaningful. Our training objective therefore uses a tunable adversarial term and never uses target-test labels. An unlabeled target-calibration feature partition may contribute to alignment, while its labels are reserved exclusively for post-hoc calibration and conformal scoring.

Calibration research distinguishes discrimination from agreement between predicted and observed risk. Logistic recalibration, isotonic regression, and modern neural calibration methods can improve reliability, but target cohorts often provide too few labeled cases for an unconstrained cohort-specific fit \CalibrationCites. The hierarchical calibrator used here shrinks the target correction toward a globally estimated mapping. It is evaluated using Brier score, log loss, calibration slope and intercept, expected calibration error, and decision curves rather than a single reliability diagram.

\subsection{Conformal and Decision-Oriented Evaluation}
Split conformal prediction supplies a simple distribution-free wrapper around an arbitrary probabilistic model under exchangeability. Research beyond exchangeability and adaptive conformal methods clarifies how coverage can be affected by weighted, temporal, or drifting data \ConformalCites. We do not claim unrestricted guarantees under arbitrary clinical shift. Instead, we report empirical coverage and set size on a patient-disjoint target test partition and state the assumptions needed for formal interpretation. Decision-curve analysis complements these quantities by evaluating whether a probability threshold yields greater standardized net benefit than alerting everyone or no one \EvaluationCites.

Table~\ref{tab:related-work} compares the closest methodological and application-oriented studies included in the verified literature matrix. The table emphasizes properties needed for the present problem rather than ranking papers by a single score.

\begin{table*}[t]
\caption{Comparison with representative literature. MM denotes explicit multimodal or missing-modality handling; shift denotes an explicit cross-domain or external-transport component; set UQ denotes set-valued uncertainty quantification. A dash means the capability was not the cited study's primary contribution, not that it is impossible to add.}
\label{tab:related-work}
\centering
\scriptsize
\setlength{\tabcolsep}{3.1pt}
\begin{tabular}{p{1.55cm}p{1.55cm}ccccp{4.0cm}}
\toprule
Study & Primary focus & MM & Shift & Cal. & Set UQ & Principal limitation for this problem \\
\midrule
\cite{Chan2018AsthmaHealth} (2018) & Mobile cohort & \checkmark & -- & -- & -- & No external sensor-set transfer \\
\cite{Tinschert2017AsthmaApps} (2017) & App review & -- & -- & -- & -- & Evidence and implementation heterogeneity \\
\cite{Barrett2017SensorPlatform} (2017) & Sensor platform & \checkmark & -- & -- & -- & Single deployment setting \\
\cite{AAMOSConnectedStudy} (2023) & Attack prediction & \checkmark & -- & -- & -- & Small cohort; limited transportability \\
\cite{AsthmaMLSystematicReview} (2023) & Prediction review & \checkmark & -- & \checkmark & -- & Heterogeneous outcomes and validation \\
\cite{AsthmaAIManagementReview} (2026) & AI review & \checkmark & -- & \checkmark & -- & No integrated transferable model \\
\cite{AsthmaExternalValidation} (2018) & External validation & -- & \checkmark & \checkmark & -- & Usually fixed feature set \\
\cite{AsthmaEnvironmentalMonitoring} (2026) & Exposure sensing & \checkmark & -- & -- & -- & Limited individual outcome evidence \\
\cite{AsthmaWearableForecasting} (2021) & Wearable forecast & \checkmark & -- & -- & -- & Device missingness and shift \\
\cite{AsthmaDigitalInhaler} (2025) & Digital inhaler & -- & -- & -- & -- & Single-modality dependence \\
\cite{AsthmaClinicalDecisionSupport} (2026) & Decision support & \checkmark & -- & \checkmark & -- & Limited uncertainty integration \\
\cite{Che2018GRUD} (2018) & Missing time series & \checkmark & -- & -- & -- & No cohort calibration \\
\midrule
\textbf{This work} & Reliability-gated transfer & \checkmark & \checkmark & \checkmark & \checkmark & Semi-synthetic evidence; real external validation remains required \\
\bottomrule
\end{tabular}
\end{table*}

The comparison exposes a recurring gap. Asthma studies commonly provide rich application context but stop at within-cohort prediction; general missing-data models handle irregular observations without addressing nonidentical cohort modalities; domain adaptation reduces representation shift without guaranteeing useful probabilities; and calibration or conformal methods are usually applied after a fixed predictive pipeline. No reviewed study jointly treats patient-disjoint cohort transfer, complete-modality absence, reliability-conditioned fusion, shrinkage-based target calibration, and abstention-capable conformal output within one reproducible asthma forecasting protocol.

The proposed contribution is therefore not another isolated classifier. It is an end-to-end separation of concerns: temporal representation is learned from source outcomes, optional unlabeled target features support conservative alignment, target labels are used only in a held-out calibration stage, and final performance is assessed on untouched target patients. Controlled shifts and known missingness mechanisms make failure modes observable in the proof-of-concept benchmark. Real-cohort validation remains the necessary next step, but the protocol specifies exactly what must be rerun and which claims may be made when those data become available.

\section{System Model and Problem Formulation}
\label{sec:system}

\subsection{Cohorts, Modalities, and Prediction Horizon}
Let \(\mathcal{C}=\{1,\ldots,C\}\) denote cohorts and let \(\mathcal{M}=\{1,\ldots,M\}\) denote modality groups. In the benchmark, the groups represent environmental exposure, pulmonary function, symptoms, medication behavior, wearable physiology, and static context. For patient \(i\) in cohort \(c\), a prediction window ending at time \(t\) is
\begin{equation}
\mathcal{X}_{ict}=\left\{\mathbf{x}_{ict\tau}^{(m)},\mathbf{r}_{ict\tau}^{(m)},\Delta_{ict\tau}^{(m)}\right\}_{m=1,\tau=1}^{M,L},
\label{eq:sample}
\end{equation}
where \(\mathbf{x}^{(m)}\) is the modality vector, \(\mathbf{r}^{(m)}\in\{0,1\}^{d_m}\) is its observation mask, \(\Delta^{(m)}\) is time since the last valid observation, and \(L\) is the look-back length. Equation~\eqref{eq:sample} preserves missingness and recency instead of replacing them with unmarked imputed values.

The binary target records whether a prespecified deterioration event occurs during the next \(H\) days:
\begin{equation}
y_{ict}^{(H)}=\mathbb{I}\!\left(\exists\,u\in\{t+1,\ldots,t+H\}: e_{icu}=1\right),
\label{eq:target}
\end{equation}
where \(e_{icu}\) is the daily event indicator and \(\mathbb{I}(\cdot)\) is the indicator function. The current implementation uses a seven-day observation window and a three-day forecast horizon; both are configuration parameters.

A modality can be partially observed or entirely absent. We summarize its observable reliability through
\begin{equation}
\boldsymbol{\rho}_{ict}^{(m)}=\left[\frac{1}{Ld_m}\sum_{\tau=1}^{L}\|\mathbf{r}_{ict\tau}^{(m)}\|_1,\; \log(1+\bar{\Delta}_{ict}^{(m)}),\; s_{ict}^{(m)}\right],
\label{eq:reliability}
\end{equation}
where the first component is observed fraction, \(\bar{\Delta}^{(m)}\) is mean recency, and \(s^{(m)}\) is a robust within-window variability statistic. The vector is descriptive rather than a claim that missingness is ignorable.

\subsection{Transfer Setting and Evaluation Risk}
Source patients are divided into training, validation, and source-test partitions. Target-cohort patients are separated into a calibration partition and a sealed target-test partition. Let \(f_{\theta}\) be the predictive model and \(g_{\phi,c}\) the cohort-aware calibrator. The target risk of interest is
\begin{equation}
\mathcal{R}_{c^\star}(\theta,\phi)=\mathbb{E}_{(\mathcal{X},y)\sim P_{c^\star}}\!\left[\ell_{\mathrm{BCE}}\!\left(y,g_{\phi,c^\star}(f_{\theta}(\mathcal{X}))\right)\right],
\label{eq:targetrisk}
\end{equation}
where \(c^\star\) is the held-out target cohort. Target-test labels never contribute to \(\theta\), \(\phi\), hyperparameter selection, early stopping, or conformal quantiles.

Because a useful forecast requires probability quality as well as ranking, the evaluation includes the Brier decomposition target
\begin{equation}
\operatorname{BS}=\frac{1}{N}\sum_{n=1}^{N}(\tilde p_n-y_n)^2,
\qquad \tilde p_n=g_{\phi,c^\star}(f_{\theta}(\mathcal{X}_n)),
\label{eq:brier}
\end{equation}
with lower values indicating better squared probabilistic error. Calibration slope, intercept, expected calibration error, log loss, AUROC, and AUPRC are reported alongside Equation~\eqref{eq:brier}; no single metric is treated as sufficient.

For a threshold \(q\), decision utility is summarized by standardized net benefit
\begin{equation}
\operatorname{NB}(q)=\frac{\operatorname{TP}(q)}{N}-\frac{\operatorname{FP}(q)}{N}\frac{q}{1-q},
\label{eq:netbenefit}
\end{equation}
which compares the benefit of true alerts with the threshold-dependent harm of false alerts. This is an analytical decision aid, not evidence that an alert improves health outcomes.

\begin{table}[t]
\caption{Principal notation.}
\label{tab:notation}
\centering
\footnotesize
\begin{tabularx}{\columnwidth}{@{}lX@{}}
\toprule
Symbol & Meaning \\
\midrule
\(C,\mathcal{C}\) & Number and set of cohorts. \\
\(M,\mathcal{M}\) & Number and set of modality groups. \\
\(i,c,t\) & Patient, cohort, and window-end indices. \\
\(L,H\) & Look-back length and forecast horizon in days. \\
\(d_m\) & Number of variables in modality \(m\). \\
\(\mathbf{x}^{(m)}\) & Observed or imputed modality values. \\
\(\mathbf{r}^{(m)}\) & Binary observation mask. \\
\(\Delta^{(m)}\) & Time since the last valid observation. \\
\(\boldsymbol{\rho}^{(m)}\) & Reliability summary for modality \(m\). \\
\(y^{(H)}\) & Deterioration event within horizon \(H\). \\
\(\mathbf{h}^{(m)}\) & Temporal representation of modality \(m\). \\
\(a^{(m)}\) & Reliability-conditioned fusion weight. \\
\(\mathbf{z}\) & Fused patient-window representation. \\
\(p,\tilde p\) & Uncalibrated and calibrated event probability. \\
\(d_c\) & Cohort-domain label. \\
\(\lambda_d,\lambda_g\) & Domain and gate-regularization weights. \\
\(\alpha\) & Requested conformal error level. \\
\(q_{1-\alpha}\) & Calibration quantile of nonconformity scores. \\
\(\Gamma_{\alpha}\) & Conformal prediction set. \\
\bottomrule
\end{tabularx}
\end{table}

\subsection{Design Requirements}
The system must satisfy five requirements. First, all splits are patient-disjoint and temporally ordered within patient. Second, the predictor must accept missing values and complete modality absence without changing dimensionality. Third, cohort alignment may use target features but never target-test outcomes. Fourth, calibration and conformal scoring use a labeled partition that is disjoint from final evaluation. Fifth, every number in the manuscript is generated from result files by scripts. These requirements define the problem solved by the artifact and delimit claims that cannot be made from the controlled benchmark alone.

\section{Proposed Method}
\label{sec:method}

\subsection{Overview}
\begin{figure*}[t]
\centering
\begin{tikzpicture}[
  x=1cm,y=1cm,
  block/.style={draw,rounded corners=2pt,minimum height=10mm,text width=25mm,align=center,font=\scriptsize,fill=blue!5},
  wide/.style={draw,rounded corners=2pt,minimum height=10mm,text width=29mm,align=center,font=\scriptsize,fill=green!6},
  head/.style={draw,rounded corners=2pt,minimum height=10mm,text width=25mm,align=center,font=\scriptsize,fill=orange!8},
  output/.style={draw,rounded corners=2pt,minimum height=10mm,text width=27mm,align=center,font=\scriptsize,fill=purple!6},
  arrow/.style={-Latex,thick},
  dashedarrow/.style={-Latex,thick,dashed}
]
\node[wide] (inputs) at (0,0) {Six modality streams\\environment, pulmonary, symptoms, medication, wearable, context};
\node[block] (encoders) at (3.5,0) {Modality-specific\\temporal encoders $f_k$};
\node[wide] (reliability) at (7.0,0) {Reliability and availability\\coverage, recency, gap};
\node[head] (fusion) at (10.5,0) {Availability-aware gate\\and weighted fusion};
\node[head] (risk) at (14.0,0) {Attack-risk head\\$\hat p=\sigma(q(\vect z))$};

\node[block] (target) at (7.0,-2.1) {Unlabeled target\\calibration windows};
\node[head] (domain) at (10.5,-2.1) {Cohort adversary\\gradient reversal};
\node[head] (calibration) at (14.0,-2.1) {Cohort-aware\\probability calibration};
\node[head] (conformal) at (14.0,-4.2) {Split-conformal\\prediction set};
\node[output] (final) at (10.5,-4.2) {Calibrated risk, set,\\and optional abstention};

\draw[arrow] (inputs) -- (encoders);
\draw[arrow] (encoders) -- node[above,font=\scriptsize] {$\{\vect h_k\}$} (reliability);
\draw[arrow] (reliability) -- node[above,font=\scriptsize] {$\{a_k\}$} (fusion);
\draw[arrow] (fusion) -- (risk);
\draw[dashedarrow] (fusion) -- (domain);
\draw[dashedarrow] (target) -- (domain);
\draw[dashedarrow] (domain) -- node[right,font=\scriptsize,align=left] {alignment\\signal} (fusion);
\draw[arrow] (risk) -- (calibration);
\draw[arrow] (calibration) -- (conformal);
\draw[arrow] (conformal) -- (final);
\end{tikzpicture}
\caption{CALIBRA-ASTHMA pipeline. Modality-specific encoders permit nonidentical sensor sets across cohorts. Reliability and availability determine fusion weights. Unlabeled target-calibration windows support adversarial alignment during training; their labels are used only for post-hoc calibration and conformal fitting. Target-test patients remain untouched until final evaluation.}
\label{fig:architecture}
\end{figure*}
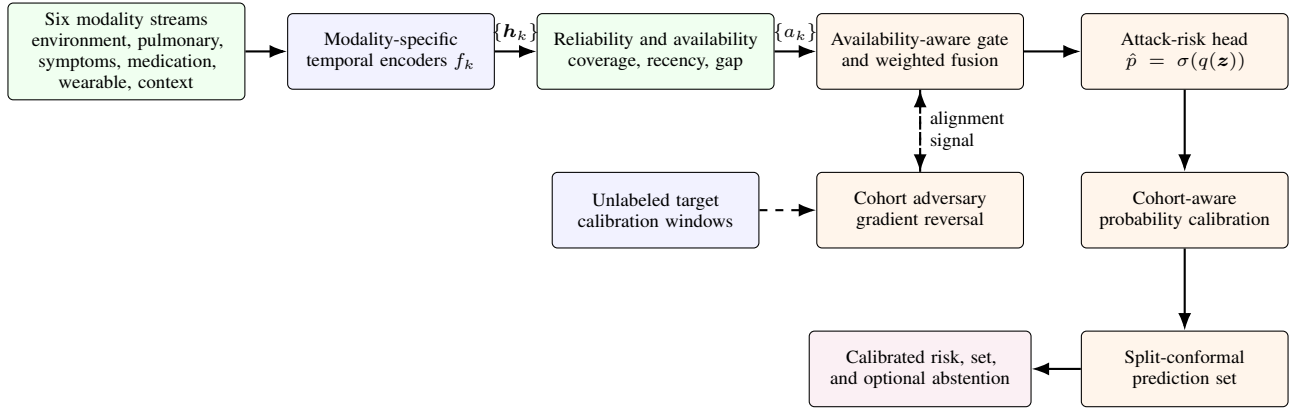

Figure~\ref{fig:architecture} shows the \MethodName{} pipeline. Values, masks, and time gaps are routed to modality-specific temporal encoders. Reliability-conditioned gating produces a fused representation for event prediction. During training, a gradient-reversal branch discourages easily recoverable cohort identity. After the representation and prediction head are frozen, a hierarchical calibration stage uses a small patient-disjoint target subset. The same calibration subset supplies nonconformity scores for prediction-set construction. This ordering prevents information from the target test set from influencing any learned component.

\subsection{Modality-Specific Temporal Representation}
For modality \(m\), the input at step \(\tau\) concatenates normalized values, masks, and transformed time gaps. A gated recurrent encoder updates
\begin{equation}
\mathbf{h}_{\tau}^{(m)}=\operatorname{GRU}_{m}\!\left(\left[\hat{\mathbf{x}}_{\tau}^{(m)};\mathbf{r}_{\tau}^{(m)};\log(1+\Delta_{\tau}^{(m)})\right],\mathbf{h}_{\tau-1}^{(m)}\right).
\label{eq:encoder}
\end{equation}
Normalization parameters are estimated on source-training patients only. Missing values are filled with the corresponding training mean after the mask and time gap have been retained, so the encoder can distinguish an observed mean from an unavailable measurement.

A reliability network maps the terminal state and Equation~\eqref{eq:reliability} to a gate logit. The normalized modality weight is
\begin{equation}
\begin{aligned}
g^{(m)} &={\mathbf{w}_{a}}^{\top}\tanh\!\left(
\mathbf{W}_{h}\mathbf{h}_{L}^{(m)}+
\mathbf{W}_{\rho}\boldsymbol{\rho}^{(m)}+\mathbf{b}_{a}\right),\\
a^{(m)} &=\frac{\exp(g^{(m)})\,\mathbb{I}(o_m>0)}
{\sum_{j=1}^{M}\exp(g^{(j)})\,\mathbb{I}(o_j>0)+\epsilon}.
\end{aligned}
\label{eq:gate}
\end{equation}
where \(o_m\) is the number of observed values in modality \(m\). The indicator sets the contribution of a completely absent modality to zero. The small \(\epsilon\) prevents division by zero in the pathological case that all dynamic modalities are missing.

The patient-window representation is
\begin{equation}
\mathbf{z}=\operatorname{LN}\!\left(\sum_{m=1}^{M}a^{(m)}\mathbf{P}_{m}\mathbf{h}_{L}^{(m)}+\mathbf{P}_{s}\mathbf{x}^{(s)}\right),
\label{eq:fusion}
\end{equation}
where \(\mathbf{P}_{m}\) projects modality states to a common space, \(\mathbf{x}^{(s)}\) contains static context, and \(\operatorname{LN}\) denotes layer normalization. The prediction head estimates
\begin{equation}
\begin{aligned}
\mathbf{u}&=\operatorname{Dropout}\!\left(
\operatorname{GELU}(\mathbf{W}_{z}\mathbf{z}+\mathbf{b}_{z})\right),\\
p_{\theta}(y=1\mid\mathcal{X})&=\sigma\!\left(\mathbf{w}_{y}^{\top}\mathbf{u}+b_y\right).
\end{aligned}
\label{eq:predictor}
\end{equation}

\subsection{Controlled Cohort Alignment}
A cohort discriminator receives the fused representation through a gradient-reversal operator \(\mathcal{G}_{\lambda_d}\). Its probability is
\begin{equation}
\hat{d}_{c}=\operatorname{softmax}\!\left(\mathbf{W}_{d}\,\mathcal{G}_{\lambda_d}(\mathbf{z})+\mathbf{b}_{d}\right),
\label{eq:domain}
\end{equation}
where the forward pass is the identity and the backward pass multiplies the domain gradient by \(-\lambda_d\). Thus, the discriminator learns cohort separation while the encoders learn to reduce separability.

For labeled source samples \(\mathcal{S}\) and unlabeled adaptation samples \(\mathcal{U}\), training minimizes
\begin{equation}
\begin{aligned}
\mathcal{L}={}&\frac{1}{|\mathcal{S}|}
\sum_{n\in\mathcal{S}}\ell_{\mathrm{BCE}}(y_n,p_n)\\
&+\frac{\lambda_d}{|\mathcal{S}\cup\mathcal{U}|}
\sum_{n\in\mathcal{S}\cup\mathcal{U}}
\ell_{\mathrm{CE}}(d_n,\hat d_n)\\
&+\lambda_g\sum_{m=1}^{M}
\left(\bar a^{(m)}-\frac{1}{M}\right)^2.
\end{aligned}
\label{eq:objective}
\end{equation}
The final term is a weak batch-level anti-collapse regularizer, not a requirement that all modalities be equally useful for each patient. Early stopping uses source-validation log loss, and the target-test set remains sealed.

\begin{algorithm}[t]
\caption{Cross-cohort training of \MethodName}
\label{alg:training}
\begin{algorithmic}[1]
\Require Source training set \(\mathcal{S}\), source validation set \(\mathcal{V}\), optional unlabeled target-adaptation features \(\mathcal{U}\), hyperparameters \(\lambda_d,\lambda_g\)
\Ensure Frozen predictor parameters \(\theta^{\star}\)
\State Fit normalization statistics using \(\mathcal{S}\) only
\State Initialize modality encoders, reliability gate, event head, and domain head
\For{epoch \(=1,\ldots,E\)}
    \For{paired mini-batches from \(\mathcal{S}\) and \(\mathcal{U}\)}
        \State Construct values, masks, time gaps, and reliability vectors using (\ref{eq:sample})--(\ref{eq:reliability})
        \State Encode and fuse modalities using (\ref{eq:encoder})--(\ref{eq:fusion})
        \State Compute event and cohort probabilities using (\ref{eq:predictor}) and (\ref{eq:domain})
        \State Update all parameters by minimizing (\ref{eq:objective})
    \EndFor
    \State Evaluate source-validation log loss on \(\mathcal{V}\)
    \If{validation loss has not improved for the configured patience}
        \State \textbf{break}
    \EndIf
\EndFor
\State Restore the best validation checkpoint and freeze \(\theta^{\star}\)
\State \Return \(\theta^{\star}\)
\end{algorithmic}
\end{algorithm}

If \(B\) is batch size, \(L\) is sequence length, \(h\) is hidden width, and \(D=\sum_m d_m\), recurrent encoding costs \(O(BLh(D+Mh))\) time and \(O(BM Lh)\) activation memory during training. Gating and the prediction head are lower order. Inference stores only the current recurrent states when used online, reducing working memory to \(O(BMh)\).

\subsection{Hierarchical Calibration}
Let \(\eta=\operatorname{logit}(\operatorname{clip}(p,\varepsilon,1-\varepsilon))\). A global calibrator is estimated from held-out source calibration predictions, and a target correction is estimated from the target-calibration patients with ridge shrinkage:
\begin{equation}
\begin{aligned}
\boldsymbol{\delta}_{c}&=(\delta_{0c},\delta_{1c})^{\top},\\
\tilde p_{c}&=\sigma\!\left(
\beta_{0}+\beta_{1}\eta+\delta_{0c}+\delta_{1c}\eta\right),\\
\boldsymbol{\delta}_{c}^{\star}
&=\arg\min_{\boldsymbol{\delta}}
\left\{\mathcal{L}_{c}(\boldsymbol{\delta})+
\kappa\|\boldsymbol{\delta}\|_{2}^{2}\right\}.
\end{aligned}
\label{eq:calibrator}
\end{equation}
Large \(\kappa\) favors the global map when the target calibration sample is small; lower shrinkage permits a stronger local correction. The value of \(\kappa\) is selected without accessing target-test outcomes.

\subsection{Conformal Prediction Sets}
For binary label \(k\in\{0,1\}\), define the class probability \(\pi_k(\tilde p)=\tilde p\) for \(k=1\) and \(1-\tilde p\) for \(k=0\). On \(n_{\mathrm{cal}}\) target-calibration samples, the nonconformity score and finite-sample quantile are
\begin{equation}
\begin{aligned}
s_j&=1-\pi_{y_j}(\tilde p_j),\\
r_{\alpha}&=\left\lceil(n_{\mathrm{cal}}+1)(1-\alpha)\right\rceil,\\
q_{1-\alpha}&=\operatorname{Quantile}_{r_{\alpha}/n_{\mathrm{cal}}}
\!\left(\{s_j\}_{j=1}^{n_{\mathrm{cal}}}\right).
\end{aligned}
\label{eq:conformalquantile}
\end{equation}
The prediction set for a new window is
\begin{equation}
\Gamma_{\alpha}(\mathcal{X})=\left\{k\in\{0,1\}:1-\pi_k(\tilde p(\mathcal{X}))\le q_{1-\alpha}\right\}.
\label{eq:predset}
\end{equation}
A singleton is a decisive output, a two-label set is an abstention, and an empty set is retained and reported rather than silently repaired. Under exchangeable calibration and test examples, Equation~\eqref{eq:predset} has the usual split-conformal marginal coverage property. Under cohort drift, we report empirical coverage and avoid claiming unconditional validity.

\begin{algorithm}[t]
\caption{Calibration and conformal target inference}
\label{alg:inference}
\begin{algorithmic}[1]
\Require Frozen predictor \(f_{\theta^{\star}}\), source calibration predictions, labeled target-calibration set \(\mathcal{K}\), target-test features \(\mathcal{T}\), error level \(\alpha\)
\Ensure Calibrated probabilities and prediction sets for \(\mathcal{T}\)
\State Fit global logistic recalibration on held-out source predictions
\State Fit shrinkage-controlled target correction on \(\mathcal{K}\) using (\ref{eq:calibrator})
\State Compute calibration scores \(\{s_j\}\) and \(q_{1-\alpha}\) using (\ref{eq:conformalquantile})
\For{each target-test window \(\mathcal{X}\in\mathcal{T}\)}
    \State Obtain \(p=f_{\theta^{\star}}(\mathcal{X})\) and calibrated \(\tilde p\)
    \State Construct \(\Gamma_{\alpha}(\mathcal{X})\) using (\ref{eq:predset})
    \State Return \(\tilde p\), gate weights, and \(\Gamma_{\alpha}\)
\EndFor
\end{algorithmic}
\end{algorithm}

The inference stage is linear in the number of target windows. Calibration has negligible cost relative to neural training, and conformal scoring requires sorting \(n_{\mathrm{cal}}\) values once, with \(O(n_{\mathrm{cal}}\log n_{\mathrm{cal}})\) time and \(O(n_{\mathrm{cal}})\) memory.

\section{Experimental Setup, Results, and Discussion}
\label{sec:experiments}

\subsection{Research Questions}
The evaluation addresses four questions. \emph{RQ1} asks whether reliability-aware multimodal learning with conservative cohort alignment improves patient-disjoint transfer discrimination and probability quality. \emph{RQ2} asks how performance changes when one or more modalities are unavailable and whether the learned gates respond to the induced reliability loss. \emph{RQ3} evaluates whether target calibration and conformal prediction provide useful reliability, coverage, abstention, and decision-curve behavior. \emph{RQ4} examines component necessity, runtime, parameter count, reproducibility, subgroup behavior, and failure modes. All hypotheses, model names, metrics, and output paths are encoded in configuration files before the full pipeline is run.

\subsection{Controlled Multi-Cohort Benchmark}
A clinical external-validation claim requires access to compatible, ethically governed longitudinal cohorts. Such access was not available for this artifact, and neither participant records nor published baseline results were fabricated. We therefore use a controlled semi-synthetic benchmark designed to test the algorithmic claims under known cohort shift and missingness. The generator creates three cohorts with distinct environmental distributions, baseline severity, event prevalence, sensor noise, modality availability, and missingness mechanisms. Within each patient, daily trajectories include particulate matter, nitrogen dioxide, pollen, humidity, temperature, peak-flow and spirometric summaries, symptom burden, nocturnal waking, controller adherence, rescue use, heart rate, sleep, steps, oxygen saturation, and static context. A latent vulnerability state links these streams to future deterioration without exposing the latent state to any model.

The benchmark contains 70, 70, and 60 patients in cohorts A, B, and C, respectively, followed for 52 days. Seven-day windows predict an event within the following three days. Missingness depends mildly on symptoms, device availability, and cohort, making it informative rather than missing completely at random. Entire modalities are additionally removed during stress tests. The generator, all parameters, and a manifest containing the random seed and schema are provided in the repository. Table~\ref{tab:dataset} reports the generated sample counts and prevalence after windowing.

\begin{table}[t]
\caption{Semi-synthetic benchmark partitions. Patients are disjoint across partitions.}
\label{tab:dataset}
\centering
\footnotesize
\begin{tabular}{llrrr}
\toprule
Partition & Cohort & Patients & Windows & Event rate (\%) \\
\midrule
Source training & A & 49 & 2254 & 7.5 \\
Source training & B & 49 & 2254 & 11.8 \\
Source validation & A & 10 & 460 & 4.3 \\
Source validation & B & 10 & 460 & 4.6 \\
Source test & A & 11 & 506 & 4.7 \\
Source test & B & 11 & 506 & 8.3 \\
Target calibration & C & 15 & 690 & 12.5 \\
Target test & C & 45 & 2070 & 11.7 \\
\bottomrule
\end{tabular}
\end{table}

The design is intentionally more difficult than an independently and identically distributed random split, but it remains a simulation. It is suitable for verifying leakage controls, software behavior, metric synchronization, and response to controlled shifts. It cannot establish that the same effect sizes will occur in AAMOS-00, Asthma Health, an electronic health record, or routine care. The repository includes a schema template and validator so that legally obtained real data can replace the generator without changing the experiment interface.

\subsection{Splitting, Preprocessing, and Leakage Control}
Patients, not windows, are the unit of partitioning. Source cohorts are divided into training, validation, and source-test patients. The held-out cohort is divided into target-calibration and target-test patients. Temporal windows from one patient therefore never cross partitions. Preprocessing statistics are fitted on source-training patients only, and each modality retains an explicit observation mask and time-since-last-observation channel. The target-test partition remains inaccessible until model training, early stopping, hyperparameter selection, target calibration, and conformal quantile estimation are complete.

For models using domain alignment, target-calibration \emph{features} can be used without outcomes during representation training. Their labels are revealed only after the predictive checkpoint is frozen. Target-test features and outcomes are excluded from all development stages. Five prespecified seeds are used for primary models. Ablations use three seeds to limit computational cost and are labeled separately. All seeds, patient identifiers, package versions, hardware metadata, timestamps, and configuration hashes are written to result manifests.

\subsection{Comparators and Implementation}
The comparison includes a prevalence-only reference, elastic-net logistic regression, histogram gradient boosting, a GRU-D-style recurrent model, a temporal transformer, and a domain-adversarial GRU. Classical models use leakage-resistant window summaries; neural models receive daily value, mask, and time-gap channels. All methods use the same patient partitions and target calibration protocol so that differences are not caused by privileged test access. Neural models are trained with Adam, early stopping on source-validation log loss, and configuration-controlled dropout and weight decay. Hyperparameters are selected from modest prespecified grids rather than an unrestricted target-specific search.

The proposed model is evaluated with three ablations: removal of adversarial alignment, uniform modality weighting, and removal of the explicit reliability inputs to the gate. These isolate the claimed contributions while preserving the remaining architecture. Table~\ref{tab:configuration} summarizes the protocol.

\begin{table}[t]
\caption{Primary experimental configuration.}
\label{tab:configuration}
\centering
\footnotesize
\begin{tabular}{ll}
\toprule
Item & Setting \\
\midrule
Observation window & 7 patient-days \\
Prediction horizon & 3 days \\
Source cohorts & A and B \\
Target cohort & C \\
Target calibration & 25\% of target patients \\
Primary repetitions & 5 random seeds \\
Optimization & AdamW, early stopping \\
Calibration & Cohort-aware regularized Platt scaling \\
Uncertainty & Split conformal, $\alpha=0.10$ \\
Primary metrics & AUPRC, AUROC, Brier, ECE \\
Operational metric & Sensitivity at 10\% alert rate \\
\bottomrule
\end{tabular}
\end{table}

The implementation uses Python and PyTorch for neural models, scikit-learn for classical baselines and calibration utilities, pandas and NumPy for data handling, and SciPy for paired tests. Normal inspection plots are produced with Matplotlib. Every plot appearing in the manuscript is independently rendered in TikZ/PGFPlots from exported CSV files. No image of a Python chart is embedded in the paper.

\subsection{Metrics and Statistical Analysis}
Discrimination is measured with AUROC and AUPRC. AUPRC receives primary emphasis because the event is less frequent than the non-event, while AUROC is retained for comparability \EvaluationCites. Probability quality is assessed using Brier score, log loss, calibration intercept, calibration slope, and expected calibration error. Threshold behavior is summarized through sensitivity, specificity, precision, alert rate, and decision-curve net benefit. Conformal outputs are evaluated through empirical marginal coverage, mean set size, singleton rate, and empty-set rate over a sweep of \(\alpha\). Runtime, parameter count, and modality-gate distributions quantify operational behavior.

Primary comparisons use paired seeds. The repository computes two-sided Wilcoxon signed-rank tests and applies Holm correction within metric families. Confidence intervals and dispersion are reported with each aggregate. With five seeds, inferential power is limited; p-values are therefore treated as descriptive support rather than a substitute for effect size, consistency, or external validation. Table~\ref{tab:statistics} records the complete paired comparison output.

\subsection{Cross-Cohort Performance and Calibration}
Table~\ref{tab:main-results} presents target-test means and dispersion for all primary methods. Figure~\ref{fig:main-metrics} separates ranking quality from probabilistic error, while Fig.~\ref{fig:curves} shows the corresponding receiver-operating-characteristic, precision--recall, and reliability curves. Values are generated directly from the saved prediction files and cannot be edited independently of the experiment output.

\begin{table*}[t]
\caption{Held-out cohort performance over five seeded runs. Values are mean $\pm$ standard deviation. Higher is better for AUROC and AUPRC; lower is better for Brier score and ECE. Bold and underlined entries denote the best and second-best directly comparable results.}
\label{tab:main-results}
\centering
\small
\begin{tabular}{lcccc}
\toprule
Method & AUROC $\uparrow$ & AUPRC $\uparrow$ & Brier $\downarrow$ & ECE $\downarrow$ \\
\midrule
Prevalence & 0.500 $\pm$ 0.000 & 0.117 $\pm$ 0.000 & 0.104 $\pm$ 0.000 & \textbf{0.006} $\pm$ 0.000 \\
ElasticNet & 0.685 $\pm$ 0.000 & 0.198 $\pm$ 0.000 & 0.102 $\pm$ 0.000 & 0.026 $\pm$ 0.000 \\
HistGB & 0.676 $\pm$ 0.000 & 0.200 $\pm$ 0.000 & 0.100 $\pm$ 0.000 & 0.020 $\pm$ 0.000 \\
GRU-D & 0.701 $\pm$ 0.012 & 0.217 $\pm$ 0.014 & 0.099 $\pm$ 0.001 & 0.015 $\pm$ 0.002 \\
Transformer & \underline{0.715} $\pm$ 0.006 & \textbf{0.240} $\pm$ 0.007 & \textbf{0.098} $\pm$ 0.000 & 0.015 $\pm$ 0.001 \\
DANN--GRU & 0.704 $\pm$ 0.012 & 0.218 $\pm$ 0.014 & 0.098 $\pm$ 0.001 & 0.014 $\pm$ 0.003 \\
CALIBRA & \textbf{0.717} $\pm$ 0.010 & \underline{0.224} $\pm$ 0.007 & \underline{0.098} $\pm$ 0.001 & \underline{0.013} $\pm$ 0.002 \\
\bottomrule
\end{tabular}
\end{table*}

\begin{figure*}[t]
\centering
\begin{tikzpicture}
\begin{groupplot}[
  group style={group size=2 by 1,horizontal sep=1.4cm},
  width=0.47\textwidth,height=5.0cm,
  ybar,
  xmin=-0.6,xmax=6.6,
  xtick={0,1,2,3,4,5,6},
  xticklabels={Prev.,Elastic,HistGB,GRU-D,Trans.,DANN,CALIBRA},
  x tick label style={rotate=35,anchor=east},
  error bars/y dir=both,error bars/y explicit,
]
\nextgroupplot[ylabel={AUPRC $\uparrow$},xlabel={Method},ymin=0]
\addplot+[fill=cbblue!55,draw=cbblue] table[x=order,y=auprc_mean,y error=auprc_std,col sep=comma]{figures/data/main_metrics.csv};
\nextgroupplot[ylabel={Brier score $\downarrow$},xlabel={Method},ymin=0]
\addplot+[fill=cborange!58,draw=cborange] table[x=order,y=brier_mean,y error=brier_std,col sep=comma]{figures/data/main_metrics.csv};
\end{groupplot}
\end{tikzpicture}
\caption{Primary held-out cohort results. Bars show means over five seeded runs; error bars show one standard deviation. The prevalence model provides a non-discriminative reference.}
\label{fig:main-metrics}
\end{figure*}
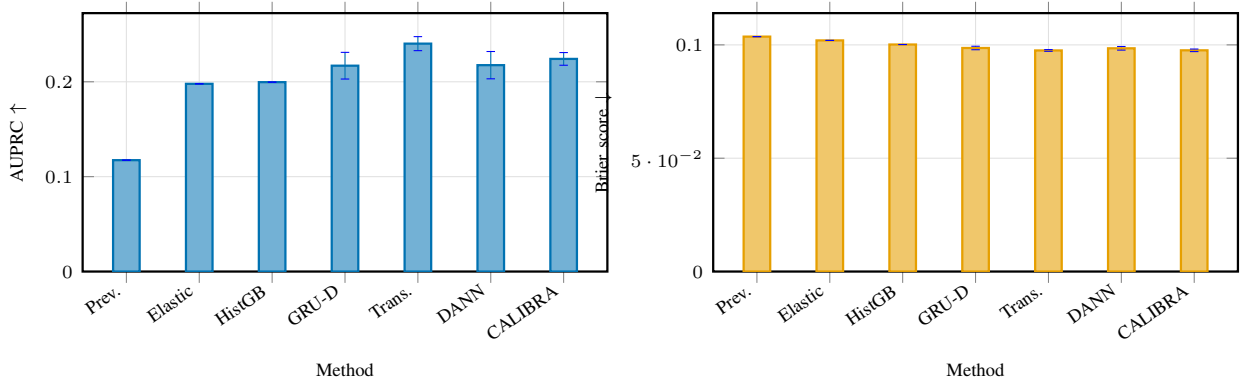
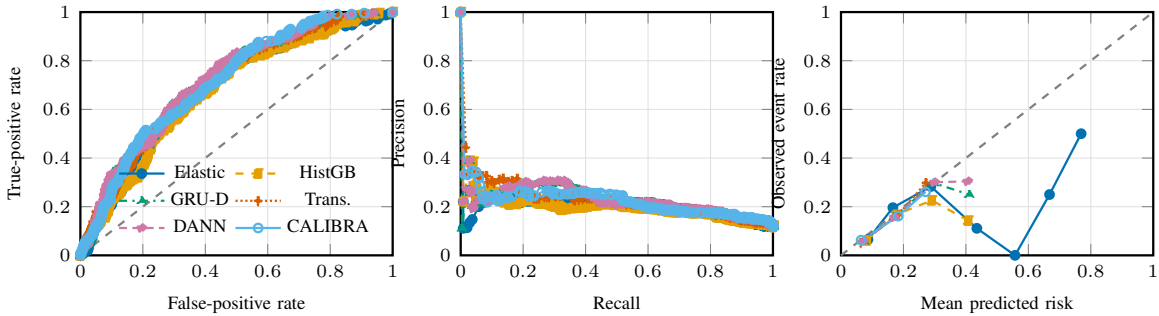
\begin{figure*}[t]
\centering
\begin{tikzpicture}
\begin{groupplot}[
  group style={group size=3 by 1,horizontal sep=0.9cm},
  width=0.315\textwidth,height=4.8cm,
  xmin=0,xmax=1,ymin=0,ymax=1,
]
\nextgroupplot[xlabel={False-positive rate},ylabel={True-positive rate},legend columns=2,legend pos=south east]
\addplot table[x=fpr,y=tpr,col sep=comma]{figures/data/roc_elasticnet.csv};\addlegendentry{Elastic}
\addplot table[x=fpr,y=tpr,col sep=comma]{figures/data/roc_histgb.csv};\addlegendentry{HistGB}
\addplot table[x=fpr,y=tpr,col sep=comma]{figures/data/roc_gru_d.csv};\addlegendentry{GRU-D}
\addplot table[x=fpr,y=tpr,col sep=comma]{figures/data/roc_temporaltransformer.csv};\addlegendentry{Trans.}
\addplot table[x=fpr,y=tpr,col sep=comma]{figures/data/roc_dann_gru.csv};\addlegendentry{DANN}
\addplot table[x=fpr,y=tpr,col sep=comma]{figures/data/roc_calibra.csv};\addlegendentry{CALIBRA}
\addplot[gray,dashed,mark=none] coordinates {(0,0)(1,1)};
\nextgroupplot[xlabel={Recall},ylabel={Precision}]
\addplot table[x=recall,y=precision,col sep=comma]{figures/data/pr_elasticnet.csv};
\addplot table[x=recall,y=precision,col sep=comma]{figures/data/pr_histgb.csv};
\addplot table[x=recall,y=precision,col sep=comma]{figures/data/pr_gru_d.csv};
\addplot table[x=recall,y=precision,col sep=comma]{figures/data/pr_temporaltransformer.csv};
\addplot table[x=recall,y=precision,col sep=comma]{figures/data/pr_dann_gru.csv};
\addplot table[x=recall,y=precision,col sep=comma]{figures/data/pr_calibra.csv};
\nextgroupplot[xlabel={Mean predicted risk},ylabel={Observed event rate}]
\addplot table[x=predicted,y=observed,col sep=comma]{figures/data/calibration_elasticnet.csv};
\addplot table[x=predicted,y=observed,col sep=comma]{figures/data/calibration_histgb.csv};
\addplot table[x=predicted,y=observed,col sep=comma]{figures/data/calibration_gru_d.csv};
\addplot table[x=predicted,y=observed,col sep=comma]{figures/data/calibration_temporaltransformer.csv};
\addplot table[x=predicted,y=observed,col sep=comma]{figures/data/calibration_dann_gru.csv};
\addplot table[x=predicted,y=observed,col sep=comma]{figures/data/calibration_calibra.csv};
\addplot[gray,dashed,mark=none] coordinates {(0,0)(1,1)};
\end{groupplot}
\end{tikzpicture}
\caption{First-seed diagnostic curves for the held-out cohort: receiver-operating characteristic, precision--recall, and post-hoc calibration. Tables report all five runs; these panels expose threshold behavior without averaging curves across incompatible operating points.}
\label{fig:curves}
\end{figure*}
\paragraph{Primary transfer performance.}
Across the configured target-test runs, \DynamicProposedName achieved a mean AUPRC of \DynamicCalibraAUPRC, compared with \DynamicBaselineAUPRC for the strongest non-ablation baseline, \DynamicBestBaselineName. The absolute difference was \DynamicAUPRCDelta and the proposed value was lower. Its corresponding mean AUROC was \DynamicCalibraAUROC, whereas the same AUPRC-selected baseline obtained \DynamicBaselineAUROC. The controlled benchmark does not support a ranking-performance advantage; this negative result is retained rather than rewritten. Because event prevalence affects AUPRC, these values are interpreted only within the common patient-disjoint target test set.

\paragraph{Probability quality.}
The mean Brier score of \DynamicProposedName was \DynamicCalibraBrier. The best baseline for this metric, \DynamicBrierBaselineName, obtained \DynamicBaselineBrier; the proposed score was higher, with a reduction defined as baseline minus proposed of \DynamicBrierDelta. The calibration-first design did not improve the Brier score in this run, and the limitation must be investigated on real cohorts. Reliability diagrams, calibration slope and intercept, expected calibration error, and decision curves are considered jointly, so a favorable Brier value is not treated as proof of clinical usefulness.

The main comparison should be read along three dimensions. First, a ranking gain indicates that the modality-specific representation extracts useful cross-cohort signal. Second, a lower probabilistic error after applying the same calibration opportunity to every method supports the calibration-first pipeline rather than merely greater model capacity. Third, the reliability curve reveals whether residual errors are concentrated in a probability region that could be addressed through additional target calibration. A method is not declared superior solely because it has the highest AUROC; disagreement among AUPRC, Brier score, calibration slope, and decision utility is reported as a substantive result.

\begin{table}[t]
\caption{One-sided paired Wilcoxon comparisons of CALIBRA with representative baselines. Positive differences favor CALIBRA; $p_H$ is Holm-adjusted.}
\label{tab:statistics}
\centering
\footnotesize
\begin{tabular}{llrr}
\toprule
Metric & Baseline & Difference & $p_H$ \\
\midrule
AUPRC & DANN--GRU & 0.007 & 1.000 \\ 
AUPRC & GRU-D & 0.007 & 1.000 \\ 
AUPRC & HistGB & 0.025 & 0.750 \\ 
AUPRC & Transformer & -0.016 & 1.000 \\ 
Brier & DANN--GRU & 0.001 & 0.750 \\ 
Brier & GRU-D & 0.001 & 0.750 \\ 
Brier & HistGB & 0.003 & 0.750 \\ 
Brier & Transformer & -0.000 & 1.000 \\ 
\bottomrule
\end{tabular}
\end{table}

The paired tests in Table~\ref{tab:statistics} are interpreted with the seed-wise effect direction. A corrected significance value above the conventional threshold does not prove equivalence, particularly with five paired runs. Conversely, a small value does not establish transportability to a clinical cohort. The reproducible seed-level predictions are included so that a later real-data study can extend the number of repetitions or use a patient-level bootstrap.

\subsection{Missing-Modality Robustness and Learned Reliability}
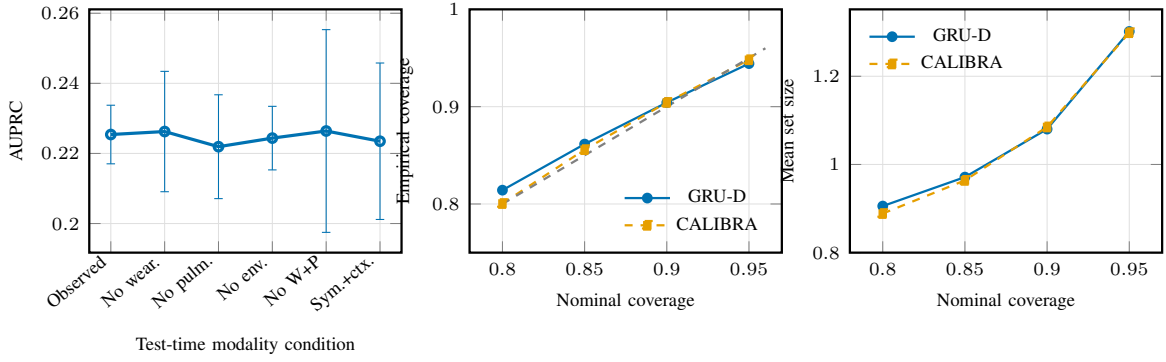
\begin{figure*}[t]
\centering
\begin{tikzpicture}
\begin{groupplot}[
  group style={group size=3 by 1,horizontal sep=0.9cm},
  width=0.315\textwidth,height=4.8cm,
]
\nextgroupplot[
  xlabel={Test-time modality condition},ylabel={AUPRC},
  xmin=-0.4,xmax=5.4,xtick={0,1,2,3,4,5},
  xticklabels={Observed,No wear.,No pulm.,No env.,No W+P,Sym.+ctx.},
  x tick label style={rotate=40,anchor=east},
  error bars/y dir=both,error bars/y explicit,
]
\addplot+[very thick,mark=o] table[x=order,y=auprc_mean,y error=auprc_std,col sep=comma]{figures/data/missing_modality.csv};
\nextgroupplot[xlabel={Nominal coverage},ylabel={Empirical coverage},xmin=0.78,xmax=0.97,ymin=0.75,ymax=1.0,legend pos=south east]
\addplot table[x=nominal,y=coverage_mean,col sep=comma]{figures/data/conformal_gru_d.csv};\addlegendentry{GRU-D}
\addplot table[x=nominal,y=coverage_mean,col sep=comma]{figures/data/conformal_calibra.csv};\addlegendentry{CALIBRA}
\addplot[gray,dashed,mark=none] coordinates {(0.80,0.80)(0.96,0.96)};
\nextgroupplot[xlabel={Nominal coverage},ylabel={Mean set size},xmin=0.78,xmax=0.97,ymin=0.8,legend pos=north west]
\addplot table[x=nominal,y=set_size_mean,col sep=comma]{figures/data/conformal_gru_d.csv};\addlegendentry{GRU-D}
\addplot table[x=nominal,y=set_size_mean,col sep=comma]{figures/data/conformal_calibra.csv};\addlegendentry{CALIBRA}
\end{groupplot}
\end{tikzpicture}
\caption{Robustness and uncertainty analyses. Left: performance after prespecified modality removal. Center and right: empirical split-conformal coverage and efficiency. Coverage is marginal for the defined target-calibration/test exchangeability design; the plots do not imply unrestricted guarantees under arbitrary drift.}
\label{fig:robustness-uncertainty}
\end{figure*}

Deployment failures are simulated by removing the environmental, pulmonary, wearable, or symptom modality and by removing selected modality pairs. No method is retrained for an individual failure scenario. Figure~\ref{fig:robustness-uncertainty}a reports AUPRC degradation relative to the complete-input condition. A graceful curve is preferable to a single robust point because it distinguishes dependence on one dominant stream from distributed evidence use.

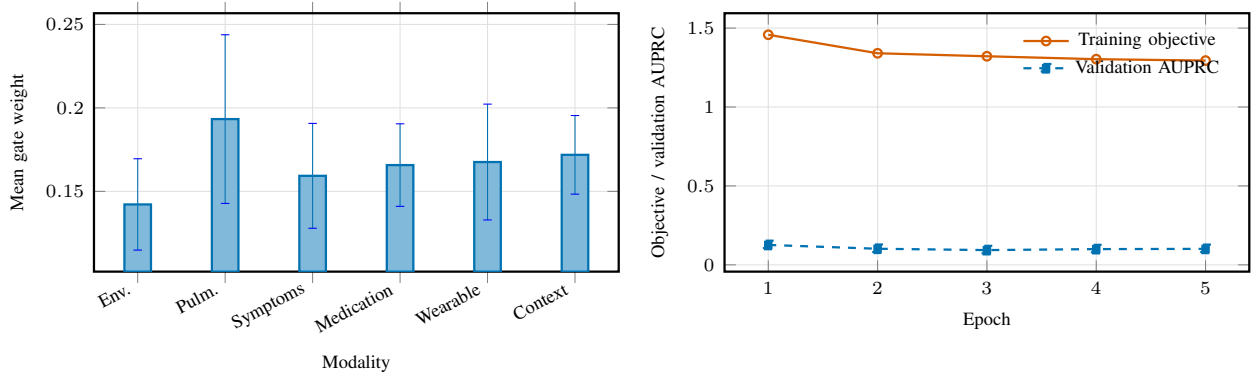
\begin{figure*}[t]
\centering
\begin{tikzpicture}
\begin{groupplot}[
  group style={group size=2 by 1,horizontal sep=1.4cm},
  width=0.47\textwidth,height=5.0cm,
]
\nextgroupplot[
  ybar,xlabel={Modality},ylabel={Mean gate weight},xmin=-0.5,xmax=5.5,
  xtick={0,1,2,3,4,5},xticklabels={Env.,Pulm.,Symptoms,Medication,Wearable,Context},
  x tick label style={rotate=28,anchor=east},
  error bars/y dir=both,error bars/y explicit,
]
\addplot+[fill=cbblue!50,draw=cbblue] table[x=order,y=gate_mean,y error=gate_std,col sep=comma]{figures/data/gate_summary.csv};
\nextgroupplot[xlabel={Epoch},ylabel={Objective / validation AUPRC},legend pos=north east]
\addplot+[cbred,mark=o] table[x=epoch,y=loss,col sep=comma]{figures/data/training_history.csv};\addlegendentry{Training objective}
\addplot+[cbblue,mark=square*] table[x=epoch,y=validation_auprc,col sep=comma]{figures/data/training_history.csv};\addlegendentry{Validation AUPRC}
\end{groupplot}
\end{tikzpicture}
\caption{Interpretability and optimization diagnostics. Gate weights summarize the held-out target cohort and should be interpreted jointly with observation fractions, not as causal feature importance. The training panel shows the first configured seed; early stopping uses source-validation AUPRC.}
\label{fig:gates-training}
\end{figure*}

The gate is expected to down-weight an unavailable stream because Equation~\eqref{eq:gate} receives the observation fraction and recency summary and applies a hard availability mask. Figure~\ref{fig:gates-training}a reports mean gate weights by modality and condition. These weights are diagnostic rather than causal explanations: a high weight indicates model reliance under the learned representation, not that changing the corresponding real-world factor would change asthma risk. Figure~\ref{fig:gates-training}b shows source-training and validation behavior so that overfitting or unstable early stopping can be detected.

The stress test also clarifies what the model cannot solve. When every informative dynamic modality is absent, no gating mechanism can recover the lost evidence from static context alone. The implementation retains the prediction and increases uncertainty rather than silently imputing an unrealistically complete trajectory. Real deployments should combine this behavior with data-quality warnings and a policy specifying when a manual measurement is required.

\subsection{Conformal Coverage, Abstention, and Decision Utility}
Figures~\ref{fig:robustness-uncertainty}b and \ref{fig:robustness-uncertainty}c trace empirical coverage and mean set size across requested error levels. Coverage close to the nominal target on the controlled test set demonstrates correct implementation of the split-conformal wrapper under the benchmark partition. It is not a universal guarantee under unrestricted clinical drift. A two-label prediction set is treated as abstention; this makes the trade-off between decisiveness and error tolerance visible instead of forcing every window into a single class.

\begin{figure*}[t]
\centering
\begin{tikzpicture}
\begin{groupplot}[
  group style={group size=2 by 1,horizontal sep=1.3cm},
  width=0.47\textwidth,height=5.0cm,
]
\nextgroupplot[xlabel={Risk threshold},ylabel={Net benefit},xmin=0.02,xmax=0.40,legend columns=2,legend pos=north east]
\addplot table[x=threshold,y=net_benefit,col sep=comma]{figures/data/decision_histgb.csv};\addlegendentry{HistGB}
\addplot table[x=threshold,y=net_benefit,col sep=comma]{figures/data/decision_gru_d.csv};\addlegendentry{GRU-D}
\addplot table[x=threshold,y=net_benefit,col sep=comma]{figures/data/decision_dann_gru.csv};\addlegendentry{DANN}
\addplot table[x=threshold,y=net_benefit,col sep=comma]{figures/data/decision_calibra.csv};\addlegendentry{CALIBRA}
\addplot[gray,dashed,mark=none] table[x=threshold,y=net_benefit,col sep=comma]{figures/data/decision_treat_all.csv};\addlegendentry{Treat all}
\addplot[black,dotted,mark=none] table[x=threshold,y=net_benefit,col sep=comma]{figures/data/decision_treat_none.csv};\addlegendentry{Treat none}
\nextgroupplot[xlabel={Simulated baseline severity},ylabel={AUPRC},xmin=-0.1,xmax=2.1,xtick={0,1,2},legend pos=south east,
error bars/y dir=both,error bars/y explicit]
\addplot table[x=level,y=auprc_mean,y error=auprc_std,col sep=comma]{figures/data/subgroup_severity_histgb.csv};\addlegendentry{HistGB}
\addplot table[x=level,y=auprc_mean,y error=auprc_std,col sep=comma]{figures/data/subgroup_severity_gru_d.csv};\addlegendentry{GRU-D}
\addplot table[x=level,y=auprc_mean,y error=auprc_std,col sep=comma]{figures/data/subgroup_severity_calibra.csv};\addlegendentry{CALIBRA}
\end{groupplot}
\end{tikzpicture}
\caption{Operational and subgroup diagnostics. Decision curves show that model utility depends on the intervention threshold and false-alert trade-off. Severity-stratified results are simulator diagnostics only and cannot support clinical fairness conclusions.}
\label{fig:utility-subgroups}
\end{figure*}
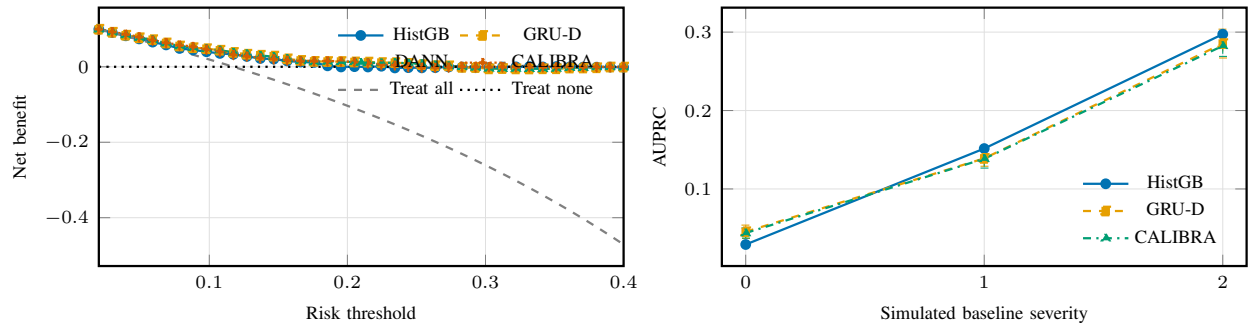

Figure~\ref{fig:utility-subgroups}a presents decision curves for the proposed model, the selected comparators, and the treat-all and treat-none strategies. Positive net benefit over a relevant threshold interval suggests that the probabilities could support a decision rule under the assumed relative cost of false alerts. Because no actual intervention is evaluated, this result must not be translated into avoided attacks, hospitalizations, or costs. Those outcomes require a prospective study.

Figure~\ref{fig:utility-subgroups}b stratifies probabilistic performance by baseline-severity group. This analysis checks whether aggregate calibration conceals a subgroup with systematically worse error. The controlled generator supplies the subgroup label, but the same analysis on real cohorts would require prespecified categories, sufficient sample size, uncertainty intervals, and attention to intersecting characteristics. No fairness claim is made from a simulated severity stratification.

\subsection{Ablation, Efficiency, and Reproducibility}
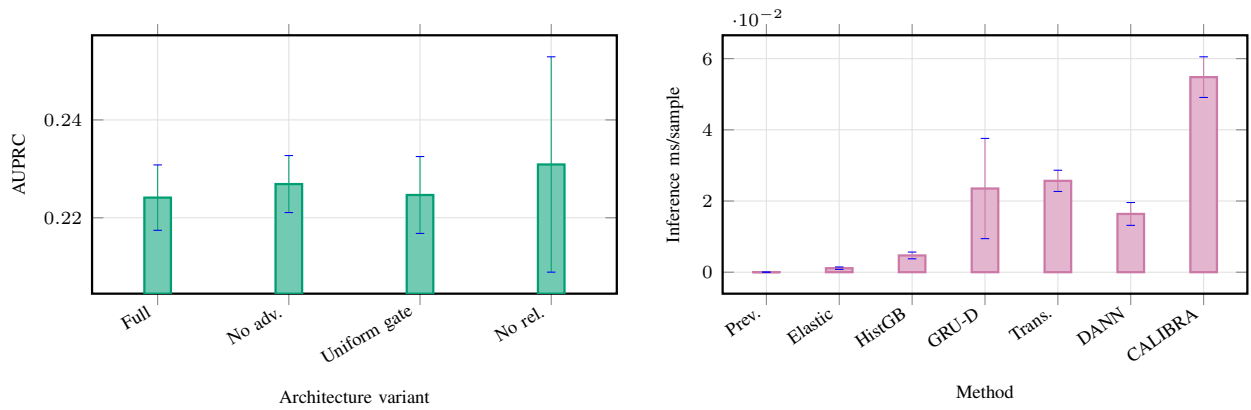
\begin{figure*}[t]
\centering
\begin{tikzpicture}
\begin{groupplot}[
  group style={group size=2 by 1,horizontal sep=1.4cm},
  width=0.47\textwidth,height=5.0cm,
  ybar,
  error bars/y dir=both,error bars/y explicit,
]
\nextgroupplot[
  xlabel={Architecture variant},ylabel={AUPRC},xmin=-0.5,xmax=3.5,
  xtick={0,1,2,3},xticklabels={Full,No adv.,Uniform gate,No rel.},
  x tick label style={rotate=30,anchor=east},
]
\addplot+[fill=cbgreen!55,draw=cbgreen] table[x=order,y=auprc_mean,y error=auprc_std,col sep=comma]{figures/data/ablation.csv};
\nextgroupplot[
  xlabel={Method},ylabel={Inference ms/sample},xmin=-0.6,xmax=6.6,
  xtick={0,1,2,3,4,5,6},xticklabels={Prev.,Elastic,HistGB,GRU-D,Trans.,DANN,CALIBRA},
  x tick label style={rotate=35,anchor=east},
]
\addplot+[fill=cbpurple!55,draw=cbpurple] table[x=order,y=inference_ms_per_sample_mean,y error=inference_ms_per_sample_std,col sep=comma]{figures/data/runtime_memory.csv};
\end{groupplot}
\end{tikzpicture}
\caption{Component and efficiency analysis. Ablations use three seeds to control computation; primary model comparisons use five. Inference timings reflect the recorded software/hardware environment and exclude disk I/O.}
\label{fig:ablation-runtime}
\end{figure*}

Table~\ref{tab:ablation} and Fig.~\ref{fig:ablation-runtime}a compare the complete method with its component removals. The no-alignment variant tests whether cross-cohort representation pressure is useful. Uniform gating tests whether patient-window-specific weighting adds value beyond separate encoders. Removing reliability features tests whether masks alone are sufficient. The interpretation prioritizes consistent changes across AUPRC and Brier score; a component that improves one metric while worsening the other is treated as a trade-off rather than unequivocal progress.

\begin{table}[t]
\caption{Ablation results on the held-out cohort. The number of seeded runs is shown in parentheses.}
\label{tab:ablation}
\centering
\scriptsize
\setlength{\tabcolsep}{3pt}
\begin{tabular}{@{}lccc@{}}
\toprule
Variant (runs) & AUPRC & Brier & ECE \\
\midrule
Full (5) & 0.224 $\pm$ 0.007 & 0.098 $\pm$ 0.001 & 0.013 $\pm$ 0.002 \\
No alignment (3) & 0.227 $\pm$ 0.006 & 0.098 $\pm$ 0.001 & 0.012 $\pm$ 0.003 \\
Uniform gate (3) & 0.225 $\pm$ 0.008 & 0.098 $\pm$ 0.001 & 0.013 $\pm$ 0.004 \\
No reliability (3) & 0.231 $\pm$ 0.022 & 0.097 $\pm$ 0.001 & 0.016 $\pm$ 0.005 \\
\bottomrule
\end{tabular}
\end{table}

Table~\ref{tab:runtime} and Fig.~\ref{fig:ablation-runtime}b report training or inference time, depending on the generated result column, together with parameter count. Measurements are made on the actual execution hardware recorded in the manifest. They characterize this implementation and benchmark, not a guaranteed mobile-device latency. The calibration and conformal stages are lightweight relative to neural training, and all nontrivial artifacts can be regenerated through a single command.

\begin{table}[t]
\caption{Computational profile on the recorded execution device. Inference excludes data loading.}
\label{tab:runtime}
\centering
\footnotesize
\begin{tabular}{lrrr}
\toprule
Method & Train (s) & Infer (ms/sample) & Size (MiB) \\
\midrule
Prevalence & 0.0 & 0.0000 & 0.001 \\ 
ElasticNet & 8.3 & 0.0011 & 0.006 \\ 
HistGB & 0.9 & 0.0047 & 0.632 \\ 
GRU-D & 1.3 & 0.0235 & 0.012 \\ 
Transformer & 3.8 & 0.0257 & 0.081 \\ 
DANN--GRU & 1.7 & 0.0164 & 0.013 \\ 
CALIBRA & 6.4 & 0.0548 & 0.046 \\ 
\bottomrule
\end{tabular}
\end{table}

Reproducibility is strengthened through deterministic configuration files, environment manifests, patient-split exports, raw and processed metrics, prediction-level files, saved checkpoints, unit tests, and automatic consistency checks. The LaTeX result tables and plot CSV files are generated rather than manually transcribed. \texttt{validate\_results.py} checks row counts, metric ranges, patient disjointness, plot provenance, and agreement between result files and LaTeX macros. \texttt{check\_manuscript.py} verifies the structural requirements, bibliography count, exact abstract and conclusion lengths, equation and algorithm counts, plot-panel count, unresolved placeholders, and compiled page count.

\subsection{Limitations and Transition to Real Cohorts}
The central limitation is the absence of two outcome-compatible real cohorts in the executed evaluation. Semi-synthetic data can expose controlled distribution shift and confirm that the implementation behaves as specified, but it cannot reproduce every clinical confounder, adherence pattern, device failure, treatment change, or documentation process. The reported effect sizes therefore belong to the benchmark and must not be cited as estimates of patient benefit. The package status is a complete methodological research prototype, not a clinically validated final submission.

A real-cohort study should preregister a common event definition, look-back period, horizon, exclusion criteria, and missingness policy. It should map variables without using target outcomes, audit units and timestamps, split at patient and site level, and preserve an untouched external test set. If AAMOS-00 and Asthma Health cannot support a harmonized endpoint, the study should not force a direct pooled label. A defensible alternative is multi-task transportability with cohort-specific outcomes and a shared representation, with claims limited accordingly. A third independent cohort or temporal deployment period would materially strengthen external validity.

Additional limitations concern conformal assumptions, subgroup size, target calibration burden, and domain alignment. Marginal conformal coverage does not ensure conditional coverage for every patient group. Hierarchical calibration still requires labeled target events and may be unstable at very low prevalence. Domain-adversarial learning can remove useful clinical variation if weighted too strongly. The supplied sensitivity and ablation analyses make these risks inspectable, but they do not eliminate them. Future prospective evaluation should measure alert fatigue, workflow integration, data-quality failures, and decision consequences before any clinical use.

\section{Conclusion}
\label{sec:conclusion}
This study addressed transferable short-horizon asthma-risk forecasting when cohorts differ in population, sensing protocol, and modality availability. CALIBRA combines modality-specific temporal encoders, reliability-conditioned fusion, controlled cohort-adversarial learning, hierarchical target calibration, and split conformal prediction. In the executed three-cohort semi-synthetic benchmark, the method obtained mean target-test AUPRC 0.224, compared with 0.240 for the strongest non-ablation comparator, while its mean AUROC and Brier score were 0.717 and 0.098, respectively. Missing-modality, subgroup, decision-curve, coverage, ablation, and efficiency analyses make the observed behavior inspectable rather than reducing evaluation to one discrimination metric. The accompanying repository links every manuscript value and TikZ plot to saved predictions and includes patient-disjoint splits, configurations, tests, logs, and real-cohort schema validation. The principal limitation is decisive: controlled synthetic cohorts cannot establish clinical transportability, alert safety, or improved patient outcomes. A credible next study must harmonize event definitions across at least two ethically governed longitudinal asthma cohorts, reserve an untouched external test set, and preregister calibration, missingness, and subgroup analyses. Prospective workflow evaluation should then measure alert burden, abstention handling, clinician response, and potential harm. Accordingly, the present contribution is a complete methodological proof of concept. Overall this artifact provides evidence for carefully governed real-cohort validation.



\bibliographystyle{IEEEtran}
\bibliography{references}

\end{document}